\documentclass{article}

\PassOptionsToPackage{numbers, sort&compress}{natbib}

\usepackage[preprint]{neurips_2023}

\usepackage[utf8]{inputenc} 
\usepackage[T1]{fontenc}    
\usepackage{hyperref}       
\usepackage{url}            
\usepackage{booktabs}       
\usepackage{amsfonts}       
\usepackage{nicefrac}       
\usepackage{microtype}      
\usepackage{xcolor}         

\usepackage{amsmath,graphicx,multirow,enumitem}
\usepackage{subcaption} 
\newcommand{\Item}[1]{\begin{itemize}#1\end{itemize}}

\newcommand{\Eqn}[1]{\begin{equation}\begin{split}#1\end{split}\end{equation}}

\newcommand{\ra}{\rightarrow}

\newcommand{\bbE}{\mathbb{E}}

\newcommand{\bbN}{\mathbb{N}}
\newcommand{\bbR}{\mathbb{R}}
\newcommand{\calC}{\mathcal{C}}
\newcommand{\calD}{\mathcal{D}}
\newcommand{\calH}{\mathcal{H}}

\newcommand{\calL}{\mathcal{L}}
\newcommand{\calM}{\mathcal{M}}
\newcommand{\calO}{\mathcal{O}}
\newcommand{\rme}{\mathrm{e}}

\newcommand{\mrm}[1]{\mathrm{#1}}

\newcommand{\hyp}{\mathrm{-}}
\newcommand{\mft}{\mathrm{MFT}}

\newcommand{\CWMS}{\mbox{CWMS}}
\newcommand{\MFCont}{\mbox{MFCont}}
\newcommand{\MFCWMS}{\mbox{MFCWMS}}

\newcommand{\better}[1]{\mathbf{#1}}
\newcommand{\best}[1]{\underline{\mathbf{#1}}}

\title{Mean Field Theory in Deep Metric Learning}

\author{%
  Takuya Furusawa \\
  ZOZO Research \\
  1-3-22 Kioicho, Chiyoda-ku, Tokyo, Japan \\
  \texttt{takuya.furusawa@zozo.com} \\
}

\begin{document}

\maketitle

\begin{abstract}
  In this paper, we explore the application of mean field theory,
  a technique from statistical physics, to deep metric learning
  and address the high training complexity
  commonly associated with conventional metric learning loss functions.
  By adapting mean field theory for deep metric learning,
  we develop an approach to design classification-based loss functions from pair-based ones,
  which can be considered complementary to the proxy-based approach.
  Applying the mean field theory to two pair-based loss functions,
  we derive two new loss functions,
  \mbox{MeanFieldContrastive} and \mbox{MeanFieldClassWiseMultiSimilarity} losses,
  with reduced training complexity.
  We extensively evaluate these derived loss functions on three image-retrieval datasets
  and demonstrate that our loss functions outperform baseline methods in two out of the three datasets.
\end{abstract}

\section{Introduction}

Deep metric learning has emerged as a powerful technique for learning meaningful data representations
in a variety of machine learning applications, such as image retrieval~\cite{wang2014learning}, face recognition~\cite{schroff2015facenet}, and person re-idenfication~\cite{hermans2017defense}.
The primary goal of deep metric learning is to provide an \textit{order} to the embedding space
by bringing similar instances closer and pushing dissimilar ones further apart.
Typically, this is achieved by optimizing a loss function that utilizes appropriate interactions based on the distance between data points.
However, conventional metric learning loss functions often suffer from high training complexity
scaling polynomially with the size of the training data.
This challenge makes optimization of these loss functions difficult in large-scale applications
and necessitates the development of sampling and mining strategies to find informative pairs

The concept of order also plays a crucial role in statistical physics,
which studies the emergent behaviors of interacting many-body systems.
These many-body systems exhibit various ordered phases of matter,
such as solid state, magnetism, superfluidity, and superconductivity,
which cannot be predicted from their individual constituents~\cite{anderson1972more}.
While the interactions between the constituents are essential for hosting such nontrivial behaviors,
they also make analyzing the systems challenging,
analogous to the issue in deep metric learning.

Mean field theory~\cite{weiss:jpa-00241247} is a powerful approach for handling the challenge associated with interacting many-body systems
and provides an insightful framework for understanding their emergent behaviors.
This theory introduces a mean field that represents the average behavior of constituent particles.
The mean field is also known as the order parameter, as its value helps distinguish the ordered phases.
The mean field theory approximates their interactions as interactions with the mean field
and significantly reduces the complexity of many-body systems.

In this paper, we leverage mean field theory from statistical physics
to tackle the complexity associated with deep metric learning.
We develop the mean field theory for applications to loss functions in deep metric learning
and find that mean field theory is better suited for losses without anchor concepts,
as opposed to the proxy-based method introduced in Ref.~\cite{movshovitz2017no}.
In this sense, it can serve as a complementary approach to the proxy-based method
for designing classification-based loss functions from pair-based ones.
Furthermore, we apply the mean field theory to two pair-based loss functions
and propose two new loss functions with reduced training complexity.
We evaluate the proposed mean field loss functions using a benchmark protocol proposed in Ref.~\cite{musgrave2020metric},
which allows us a fair comparison with other baseline loss functions,
and also using the traditional protocol utilized in Ref.~\cite{movshovitz2017no,kim2020proxy}.
The evaluation results indicate that our mean field losses surpass other methods in two out of three image-retrieval datasets in the former protocol.
Moreover, the latter evaluation protocol demonstrates
that our losses not only exhibit performance improvements
in three out of four image-retrieval datasets
but also converge more rapidly compared to \mbox{ProxyAnchor} loss~\cite{kim2020proxy}.

The main contributions of this paper are three-fold:
\textbf{(1)} the introduction of the mean field theory from statistical physics as a tool to reduce the training complexity of pair-based loss functions
based on an analogy between magnetism and deep metric learning;
\textbf{(2)} the derivation of \mbox{MeanFieldContrastive} and \mbox{MeanFieldClassWiseMultiSimilarity} losses
by application of the mean field theory to two pair-based loss functions;
\textbf{(3)} the demonstration that the derived loss functions are competitive with existing baseline losses in several datasets.

\section{Related work}

\subsection{Pair-based loss functions}

Pair-based loss functions, a representative category of deep metric learning losses,
leverage pairwise or triplet relationships among data points.
Contrastive loss~\cite{hadsell2006dimensionality} is an early example of this type,
which utilizes positive and negative pairs of data
and learns embeddings to place the positive pairs close and negative ones apart.
Triplet loss~\cite{weinberger2009distance} is an extension of Contrastive loss,
which exploits triplets of positive, negative, and anchor data,
and places an anchor embedding closer to a positive than a negative.
These losses are further extended to incorporate interactions among pairs in mini-batches to improve performance and convergence speed~\cite{sohn2016improved,oh2016deep,wang2019ranked,wang2019multi}.
For instance, MultiSimilarity loss~\cite{wang2019multi} takes into account multiple inter-pair relationships within a mini-batch, enabling more efficient learning of the embedding space.

However, a major drawback of the pair-based losses is their sensitivity to the choice of positive and negative pairs,
which is caused by the polynomial growth of the number of pairs and triplets with respect to the number of training data~\cite{schroff2015facenet}.
This usually requires sophisticated sampling and mining strategies for informative pairs
to improve performance and mitigate slow convergence~\cite{schroff2015facenet,shi2016embedding,hermans2017defense,wu2017sampling,yuan2017hard,harwood2017smart,wang2019multi}.
In this paper, we pursue an alternative approach to reduce training complexity rather than investigate these strategies.
We accomplish this by leveraging the mean field theory, a concept from statistical physics.

\subsection{Classfication-based loss functions}
Classification-based loss functions utilize weight matrices and learn embeddings by optimizing a classification objective.
Unlike pair-based losses, these losses do not face the complexity issue because they are computed in the same manner as a typical classification task.

A representative example is NormalizedSoftmax loss~\cite{wang2017normface,zhai2018classification},
which is obtained by the cross-entropy loss function with an L2-normalized weight matrix.
Its extensions include SphereFace~\cite{liu2017sphereface}, ArcFace~\cite{deng2019arcface}, and CosFace~\cite{wang2018additive,wang2018cosface} losses obtained by modifying distance metrics and introducing margins.
The losses with proxies such as ProxyTriplet and ProxyNCA losses also belong to this category~\cite{movshovitz2017no}.
Such proxy losses can be derived from corresponding pair-based losses by substituting positive and negative data points with learnable embeddings called proxies while retaining anchors.
ProxyAnchor loss~\cite{kim2020proxy} further considers interactions among samples in a mini-batch and shows promising performance in popular public datasets, surpassing other classification-based and pair-based loss functions.
These losses have been extended to incorporate refined structures among data,
such as graphs~\cite{Zhu2020fewer}, hierarchies~\cite{yang2022hierarchical}, and others~\cite{qian2019softtriple,teh2020proxynca++,li2022informative}.

In this paper, we develop the mean field theory as a technique to derive a classification-based loss from a pair-based one,
addressing the challenges of the latter.
Although our approach is similar to the proxy-based method, it naturally adapts to pair-based losses without anchors,
which have remained unexplored by the proxy-based method.

\section{Proposed approach}

In this section, we investigate the mean field theory and its application to deep metric learning.
We first review the mean field theory for a ferromagnet in statistical mechanics
by following standard statistical mechanics textbooks (e.g., see Ref.~\cite{nishimori2010elements}).
Next, based on the analogy between the ferromagnet and deep metric learning,
we apply the mean field approximation to the Contrastive loss and a variant of the MultiSimilarity loss
and derive classification-type loss functions with reduced training complexity.

\subsection{Mean field theory for magnets}\label{sec: mft in magnet}
Magnetism is one of the representative phenomena in statistical physics
that show a phase transition between ordered and disordered phases.
A ferromagnet is composed of a large number of microscopic magnetic spins
and shows a macroscopic magnetization
when a macroscopic number of the magnetic spins are aligned in the same direction.

To explain the mean field theory for a ferromagnet,
we shall consider an infinite-range model\footnote{
  Note that readers might worry that this model appears too simple
  (e.g., it lacks a notion of lattice structure).
  However, it is sufficient for explaining the phase transition of ferromagnets
  and is analogous to loss functions in deep metric learning.
} whose Hamiltonian (or energy) takes the following form~\cite{nishimori2010elements}:
\Eqn{ \label{eq: hamiltonian}
\calH = -\frac{J}{2} \sum^N_{i,j=1} S^T_i \cdot S_j,
}
with the total number of spins $N \in \bbN$ and the exchange interaction $J>0$.
Here, $S_i \in S^2$ represents the $i$-th constituent magnetic spin.
Eq.\eqref{eq: hamiltonian} indicates that a state where spins point in the same direction is preferred energetically.

According to statistical mechanics, a probabilistic distribution of the spin configuration at temperature $T$ follows the Gibbs distribution:
\Eqn{ \label{eq: partition function}
  P(\{ S_i \}_i) = \frac{\rme^{-\calH/T}}{Z}, \quad Z = \int \prod_i d^2S_i \rme^{-\calH/T},
}
and macroscopic properties of this system can be computed from the normalization factor $Z$.
However, since the spins are interacting with each other,
it is not easy to compute $Z$ both analytically and numerically.

To address this difficulty, we introduce the mean field theory.
The central idea of the theory is to approximate the Hamiltonian~\eqref{eq: hamiltonian}
such that each spin interacts with an average field generated by the rest of the spins,
rather than with other spins directly, thereby ignoring their fluctuations.
More concretely, it means that we expand $\calH$
with respect to fluctuations $\{(S_i -M)\}_i$
using the identity, $S_i = M + (S_i -M)$
and ignore the second-order terms of the expansion.
This operation results in
\Eqn{ \label{eq: mf hamiltonian}
\calH \simeq \calH_\mft =
\frac{JN^2}{2} M^T \cdot M - JNM^T \cdot \sum^N_{i=1} S_i.
}
Since $\calH_\mft$ does not include interaction terms between spins,
one can readily compute any information from the Gibbs distribution for this Hamiltonian as follows:
\Eqn{
  \label{eq: mf partition function}
  P_\mft(\{ S_i \}_i) = \frac{\rme^{-\calH_\mft/T}}{Z_\mft}, \quad Z_\mft = \int \prod_i d^2S_i \rme^{-\calH_\mft/T}.
}

The value of the mean field $M$ must be determined to minimize $-\log Z_\mft$.
This condition is justified because we can show that it is equivalent to
\Eqn{ \label{eq: consistent eq}
  M = \frac{1}{N}\sum_i\bbE[S_i]
}
by differentiating $-\log Z_\mft$ with respect to the mean field.
Here, we take the expectation value over the Gibbs distribution $P_\mft$ in Eq.~\eqref{eq: consistent eq}.
Since the mean field approximation is based on the expansion with respect to the fluctuations around the mean field,
Eq.~\eqref{eq: consistent eq} ensures the consistency of expansion.

Overall, the mean field theory is a powerful tool that allows us
to describe and analyze complex systems by approximating the interactions
between individual constituents with an average field generated by the rest of the system.
To draw a parallel between the above discussion and deep metric learning,
let us consider the $T\ra0$ limit.
In this limit, the original problem of computing $Z$ becomes
one of finding a spin configuration that minimizes $\calH$.
This is analogous to a machine learning problem
that seeks optimal parameters to minimize a loss function.
Then, the mean field approximation reduces the problem
to one of minimizing $\calH_\mft$ with respect to both the spins and mean field.
Therefore, this observation indicates that applications of mean field theory to deep metric learning problems introduce
mean fields as parameters learned to minimize their loss functions.

\subsection{Mean field contrastive loss}\label{sec: mfcont}

To study how the mean field theory works for loss functions in deep metric learning,
let us begin by applying the mean field theory to Contrastive loss for the sake of simplicity
and then proceed to discuss the mean field theory for a more complicated loss function.

In the following sections, we denote training data by $\calD = \{ x_i, y_i\}^{|\calD|}_{i=1}$ composed of input data $x_i$ and its class label $y_i \in \calC = \{1,\cdots,|\calC|\}$.
We also denote a set of data in class $c\in \calC$ as $\calD_{c}$.
We extract features from the input data using a machine learning model $F_\theta$,
whose learnable parameters are represented by $\theta$.
This model embeds the input into a manifold $\calM$, such as $\bbR^d$ or $S^d$, with $d \in \bbN$.
We also define the distance between two embeddings, $F, F' \in \calM$, as $d(F, F')\ge0$.
For instance, the distance can be given by the cosine distance for $\calM=S^d$,
taking the form $d(F, F') = 1-F^T\cdot F'/(||F||_2||F'||_2)$,
or by the Euclidean distance for $\calM=\bbR^d$.

Contrastive loss is one of the primitive examples in deep metric learning, which is defined as
\Eqn{\label{eq: cont}
  \calL_\mrm{Cont.}
  =& \frac{1}{2|\calC|} \sum_{c \in \calC} \frac{1}{|\calD_c|^2}\sum_{i,j\in \calD_c}
  \Big[ d(F_\theta(x_i),F_\theta(x_j)) - m_\mrm{P} \Big]_+
  \\
  &+ \frac{1}{2|\calC|}\sum_{c \neq c'}\frac{1}{|\calD_c||\calD_{c'}|} \sum_{i\in \calD_c, j\in \calD_{c'}}
  \Big[m_\mrm{N} - d(F_\theta(x_i),F_\theta(x_j)\Big]_+ ,
}
with $[x]_+ = \max(x,0)$.
Here, $m_\mrm{P}$ ($m_\mrm{N}$) is a hyperparamter that controls distances between positive (negative) instances.
Note that Eq.~\eqref{eq: cont} reduces to the Hamiltonian~\eqref{eq: hamiltonian}
when $|\calC|=1$ and $m_\mrm{P}<0$, and it requires the $\calO(|\calD|^2)$ training complexity
as it is parallel to the situation in Sec.~\ref{sec: mft in magnet}.

This analogy encourages us to apply the mean field theory in order to obtain a simpler loss function.
Since we have multiple classes here, we shall introduce mean fields $\{M_c\}_{c\in\calC}$ and expand $\calL_\mrm{Cont.}$ with respect to fluctuations around them.
Note that, in contrast to the single-class case, we must impose the following conditions to constrain relative distance among the mean fields:
\Eqn{\label{eq: cont constraint}
\Big[m_\mrm{N} - d(M_c,M_{c'})\Big]_+ = 0 \quad \Bigl( c\neq c' \Bigr).
}
This condition means that we should explore configurations of the mean fields
which minimize $\calL_\mrm{Cont.}$ at the zeroth order of expansions around the mean fields.
In practice, we take into account these constraints softly.

In the expansion around the mean fields,
we ignore all cross-product terms of the fluctuations keeping any others
so that we reduce the complexity while taking into account the higher-order terms of self-interactions.
By summing over the remaining terms, we obtain \mbox{MeanFieldContrastive} (MFCont.) loss,
which takes the following form:
\Eqn{\label{eq: mf cont regit}
\calL_\mrm{MFCont.} =& \frac{1}{|\calC|} \sum_{c \in \calC} \frac{1}{|\calD_c|}\sum_{i\in \calD_c}
\Big(
\Big[d(F_\theta(x_i),M_c) - m_\mrm{P}\Big]_+ + \sum_{c'(\neq c)}\Big[m_\mrm{N} - d(F_\theta(x_i),M_c)\Big]_+
\Big)
\\
& + \frac{\lambda_\mrm{MF}}{|\calC|} \sum_{c \neq c'}
\Big[m_\mrm{N}-d(M_c,M_{c'})\Big]^2_+,
}
where we impose the constraints~\eqref{eq: cont constraint} softly by $\lambda_\mrm{MF}>0$.
Note that resummation here na\"ively produces unstable terms, $\{- [m_\mrm{N} - (d(M_c,M_{c'})]_+\}_{c,c'}$,
but they vanish, thanks to the constraints~\eqref{eq: cont constraint}.
We emphasize that we must minimize $\calL_\mrm{MFCont.}$ by optimizing both $M_c$ and $\theta$,
and we can readily show that the optimal mean fields satisfy $M_c = \sum_{i \in \calD_c} F_\theta(x_i)/|\calD_c|$ at the first order of fluctuations.

\subsection{Mean field class-wise multisimilarity loss}\label{sec: mfcwms}

Lastly, we consider the mean field approximation of a loss function
which incorporates interactions within a mini-batch similar to MultiSimilarity~\cite{wang2019multi} and \mbox{ProxyAnchor}~\cite{kim2020proxy} losses.
However, the mean field approximation relies on expansions around mean fields,
and thus, a loss function symmetric with respect to $x_i$ and $x_j$ (i.e., without anchors) would be more desirable for our purpose.
Since most loss functions do not exhibit such a symmetric property,
we propose the following loss function that satisfies these requirements:
\Eqn{\label{eq: cwms}
  \calL_\mrm{CWMS} =& \frac{1}{\alpha |\calC|} \sum_{c \in \calC} \log\left[
    1+ \frac{\sum_{i,j \in \calD_c} e^{\alpha ( d(F_\theta(x_i),F_\theta(x_j)) - \delta)}}{2|\calD_c|^2}
    \right]\\
  &+ \frac{1}{2\beta |\calC|} \sum_{c \neq c'} \log\left[
    1+ \frac{\sum_{i \in \calD_c, j \in \calD_{c'}} e^{ - \beta ( d(F_\theta(x_i),F_\theta(x_j)) - \delta)}}{|\calD_c||\calD_{c'}|}
    \right],
}
with hyperparameters $\alpha$, $\beta$, and $\delta$.
Since Eq.~\eqref{eq: cwms} takes a similar form to MultiSimilarity loss~\cite{wang2019multi}
but incorporates interactions among negative samples in a class-wise manner,
we refer to it as \mbox{ClassWiseMultiSimilarity} (CWMS) loss.

Next, we derive the mean field counterpart of this loss function.
Here, the logits in the first and second terms take forms
similar to the positive and negative interactions found in Contrastive loss~\eqref{eq: cont}.
Repeating the discussion in Sec.~\ref{sec: mfcont}, we derive \mbox{MeanFieldClassWiseMultiSimilarity} (MFCWMS) loss:
\Eqn{\label{eq: mfcwms}
\calL_\mrm{MFCWMS} =& \frac{1}{\alpha |\calC|} \sum_{c \in \calC} \log\left[
  1 + \frac{\sum_{i\in \calD_c} e^{\alpha ( d(F_\theta(x_i),M_c) - \delta)}}{|\calD_c|}
  \right]\\
+& \frac{1}{2\beta |\calC|} \sum_{c \neq c'} \log\left[
1+ \frac{\sum_{i \in \calD_c} e^{ - \beta ( d(F_\theta(x_i),M_{c'}) - \delta)}}{|\calD_c|}
+ \frac{\sum_{j \in \calD_{c'}} e^{ - \beta ( d(M_c,F_\theta(x_j)) - \delta)}}{|\calD_{c'}|}
\right]
\\
+& \frac{\lambda_\mrm{MF}}{|\calC|} \sum_{c \neq c'} \left(\log\left[
1+ e^{ - \beta ( d(M_c,M_{c'}) - \delta)}
\right]\right)^2,
}
where we also introduce the soft constraint for the mean fields to ignore unstable terms produced in the resummation.\footnote{
Rigorously speaking, we cannot minimize $\{e^{ - \beta ( d(M_c,M_{c'}) - \delta)}\}_{c,c'}$ simultaneously.
However, focusing on the region with $\beta\gg1$,
we can easily find mean field configurations
satisfying $e^{ - \beta ( d(M_c, M_{c'}) - \delta)}\ll1$ at the same time.
This is enough to ignore unstable terms in practice.
}
Compared to \mbox{MeanFieldContrastive} loss,
this loss function incorporates interactions of positive samples
as well as those of negative ones in a class-wise manner like \mbox{ProxyAnchor} loss.
\begin{figure}[t]
  \centering
  \includegraphics[scale=0.7]{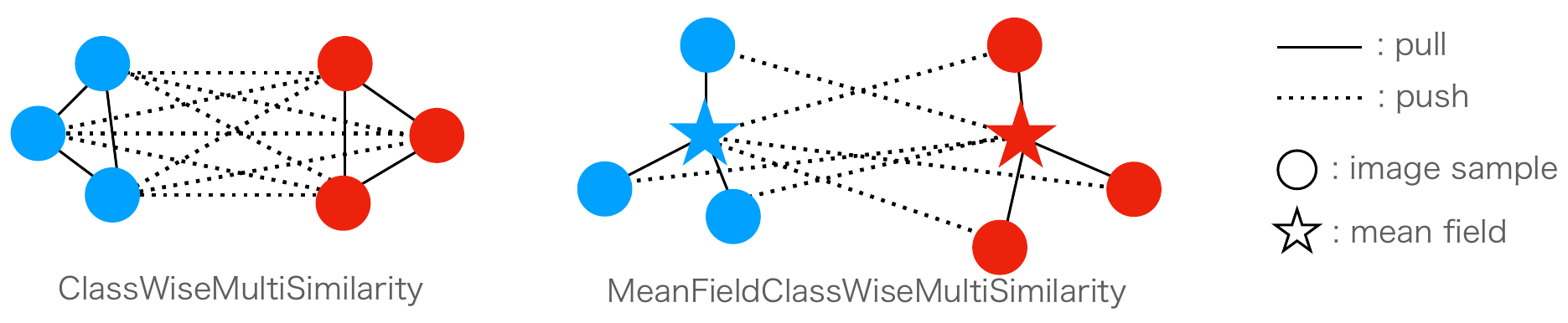}
  \caption{
    Schematic illustration of interactions in \mbox{CWMS} (left) and \mbox{MFCWMS} (right) losses.
    Each color indicates a class to which an embedding and a mean field belong.}
  \label{fig: cwms}
\end{figure}

\section{Experiments}
Let us see the effectiveness of the proposed mean field losses by evaluating their image-retrieval performance in several public datasets.
We employ the recently proposed benchmarking scheme~\cite{musgrave2020metric}
as well as the traditional one used in Refs.~\cite{movshovitz2017no,kim2020proxy}.
We compare our mean field losses and existing loss functions,
such as MultiSimilarity and \mbox{ProxyAnchor} losses.
We also explore the effect of hyperparameters on their evaluation metrics.
In our experiment, we use precision at $1$ (P@1), R-precision (RP), and mean average precision at R (MAP@R) as evaluation metrics.
In particular, we focus on MAP@R because it reflects the correctness of the ranking for retrievals
and is a suitable metric to evaluate the quality of the embedding space~\cite{musgrave2020metric}.
Note that we implement our experiments in PyTorch~\cite{paszke2019pytorch}
and utilize the PyTorch Metric Learning library~\cite{Musgrave2020PyTorchML} to implement baseline models.

\subsection{Datasets}\label{sec: datasets}

In our experiments, we utilize four publicly available image-retrieval datasets,
CUB-200-2011 (CUB)~\cite{wah2011caltech}, Cars-196 (Cars)~\cite{krause20133d},
Stanford Online Products (SOP)~\cite{oh2016deep}, and In-Shop~\cite{liu2016deepfashion}.
CUB comprises $11{,}788$ images of birds categorized into $200$ classes,
while the Cars dataset consists of $16{,}185$ images of $196$ car classes.
In CUB, the first $100$ classes ($5{,}864$ images) are used for the training dataset,
and the remaining $100$ classes ($5{,}924$ images) are allocated for the test dataset.
Similarly, Cars was split into $8{,}054$ training images ($98$ classes) and $8{,}131$ test images ($98$ classes).
The SOP dataset contains $22{,}634$ classes with $120{,}053$ product images.
The initial $11{,}318$ classes with $59{,}551$ images are used for training,
and the remaining $11{,}316$ classes with $60{,}502$ images are allocated for testing.
Lastly, the InShop dataset features $52{,}712$ images of $7{,}982$ fashion products,
with $25{,}882$ images from $3{,}997$ classes used for training and $26{,}830$ images from $3{,}985$ classes allocated for testing,
which are further divided into query ($14{,}218$ images) and gallery ($12{,}612$ images) subsets.

\subsection{Implementation details}

As a backbone embedding model $F_\theta(x)$,
we employ the inception network with batch normalization (BN-Inception)~\cite{ioffe2015batch},
which is pretrained for the classification task on the ImageNet dataset~\cite{russakovsky2015imagenet}.
We reduce the embedding dimensions
by inserting a fully-connected layer with ReLU activation functions in the first scheme
and replacing its last linear layer with that of desired dimensions in the second scheme.
In both cases, we apply random resized cropping and random horizontal flipping to all inputs during training
and only center cropping during evaluation.

In the modern benchmarking protocol,
we perform $50$ iterations of Bayesian optimization for hyperparameters in loss functions
including the learning rate for proxies and mean fields for a fair comparison.
We split a dataset into train--valid (the first half classes) and test datasets (the remainings).
The train--valid set was further divided into four partitions in a class disjoint manner,
and we performed four-fold cross-validation based on the leave-one-out method in each iteration.
In each cross-validation step, we train a model with embedding dimensions set to $128$ and batch size set to $32$
until MAP@R for the validation data converges.
The Bayesian optimization aims to maximize the average of the four validation metrics.
Note that we sample images so that each mini-batch is composed of $32$ classes ($8$ classes) and $1$ image ($4$ images) per class
for classification-based (pair-based) losses,
and we utilize the RMSprop optimizer with learning rate $10^{-6}$ for the embedding model.
In the test stage, we perform cross-validation again with the best hyperparameters, resulting in four embedding models.
Using these models, we evaluate performance on the test dataset in the following two different ways:
mean of the metrics computed from the $128$-dimensional ($128$D) embeddings (separated)
and those from $512$D embeddings made of the four $128$D ones (concatenated).
We repeat this observation $10$ times and report their average values with $95\%$ intervals.
We carry out the experiments on a single NVIDIA V100 GPU.

In contrast, in the traditional evaluation protocol,
we use the predefined train--test splits described in Sec.~\ref{sec: datasets}
and train a model for up to $60$ epochs with embedding dimensions $512$ and batch size $128$,
setting the patience for early stopping to $5$ to accelerate the experiments.
In this case, we use AdamW optimizer~\cite{loshchilov2017decoupled}
with the learning rate $10^{-4}$ for the embedding model,
setting the learning rate for proxies to $10^{-2}$ and that for mean fields to $2\times10^{-1}$.
The hyperparameters for \mbox{ProxyAnchor} loss are fixed to $(\alpha,\delta)=(32,10^{-1})$,
while we set $(m_\mrm{P},m_\mrm{N},\lambda_\mrm{MF}) = (0.02,0.3,0)$ for \mbox{MFCont}. loss
and $(\alpha,\beta,\delta,\lambda_\mrm{MF})=(0.01,80,0.8,0)$ for \mbox{MFCWMS} loss in default.
We chose these default parameters according to the results of the Bayesian optimization
and the discussion in Sec.~\ref{sec: hyp}.
We repeat the above procedure $10$ times and report the averages of the metrics
computed from the test embedding with the best MAP@R with $95$\% confidence intervals.
The experiments on the CUB and Cars datasets are carried out on a single NVIDIA V100 GPU,
while those on the SOP and InShop datasets are performed on a single NVIDIA A100 GPU.

\subsection{Benchmark results}

Based on the first protocol, we study the performance of our loss functions on three datasets; CUB, Cars, and SOP.
We compare our loss functions with existing ones,
such as Contrastive, MultiSimilarity, ArcFace, CosFace, ProxyNCA, and \mbox{ProxyAnchor} losses.
The experimental results are summarized in Table~\ref{tbl: mlrc} (see the supplement for complete results).
First, the mean field losses show better performance than their original pair-based losses in most cases,
indicating that applying mean field theory not only reduces training complexity but also results in better embeddings.
Furthermore, the mean field losses consistently outperform other baseline methods in both separate and concatenated MAP@R
for the CUB and SOP datasets.
However, in Cars, \mbox{ProxyAnchor} and \mbox{CWMS} losses show better performance than the mean field losses,
which might imply the importance of interactions within batch samples in this dataset.

\tabcolsep=0.15cm
\begin{table}[t]
  \caption{
    MAP@R obtained from the modern protocol~\cite{musgrave2020metric} in CUB-200-2011, Cars-196, and Stanford Online Products.
    We carry out test runs $10$ times and present the averaged metrics along with their confidence intervals.
    The best result within each block is highlighted in bold,
    while the overall best results for all losses are underlined.
    \mbox{ProxyAnchor} loss failed to converge in the SOP dataset in our settings.
    See the supplement for complete results.
  }
  \label{tbl: mlrc}
  \centering
  \begin{tabular}{ccccccc}
    \toprule
               & \multicolumn{2}{c}{CUB} & \multicolumn{2}{c}{Cars} & \multicolumn{2}{c}{SOP}                                                                         \\
    \cmidrule(r){2-3}\cmidrule(r){4-5}\cmidrule(r){6-7}
    Loss       & $128$D                  & $512$D                   & $128$D                  & $512$D                & $128$D                & $512$D                \\
    \midrule
    ArcFace    & ${21.5\pm0.1}$          & ${26.4\pm0.2}$           & ${18.3\pm0.1}$          & $\better{27.6\pm0.1}$ & ${41.5\pm0.2}$        & $\better{47.4\pm0.2}$ \\
    CosFace    & ${21.2\pm0.2}$          & $\better{26.5\pm0.3}$    & ${18.5\pm0.1}$          & ${27.0\pm0.3}$        & ${41.0\pm0.2}$        & ${46.8\pm0.2}$        \\
    MS         & ${21.0\pm0.2}$          & ${26.2\pm0.2}$           & ${18.7\pm0.3}$          & ${27.2\pm0.4}$        & ${41.9\pm0.2}$        & ${46.7\pm0.2}$        \\
    MS+Miner   & ${20.8\pm0.2}$          & ${25.9\pm0.2}$           & ${18.5\pm0.2}$          & ${26.9\pm0.4}$        & ${41.9\pm0.3}$        & ${46.6\pm0.3}$        \\
    ProxyNCA   & ${18.8\pm0.2}$          & ${23.8\pm0.2}$           & ${17.4\pm0.1}$          & ${26.8\pm0.2}$        & $\better{42.7\pm0.1}$ & ${46.7\pm0.1}$        \\
    ProxyAnch. & $\better{21.7\pm0.2}$   & $\better{26.5\pm0.2}$    & $\best{19.4\pm0.2}$     & ${26.8\pm0.3}$        & $\hyp$                & $\hyp$                \\
    \midrule
    Cont.      & ${21.0\pm0.1}$          & ${26.4\pm0.2}$           & ${17.0\pm0.3}$          & ${24.9\pm0.5}$        & ${41.1\pm0.2}$        & ${45.3\pm0.2}$        \\
    \MFCont.   & $\better{22.0\pm0.1}$   & $\best{27.2\pm0.1}$      & $\better{18.1\pm0.1}$   & $\better{27.4\pm0.2}$ & $\better{43.6\pm0.4}$ & $\better{47.0\pm0.2}$ \\
    \midrule
    \CWMS      & ${21.5\pm0.3}$          & ${26.9\pm0.3}$           & $\better{19.3\pm0.3}$   & $\best{27.8\pm0.3}$   & ${41.5\pm0.2}$        & ${45.1\pm0.2}$        \\
    \MFCWMS    & $\best{22.1\pm0.1}$     & $\better{27.0\pm0.1}$    & ${18.9\pm0.2}$          & ${27.0\pm0.3}$        & $\best{44.6\pm0.2}$   & $\best{48.3\pm0.2}$   \\
    \bottomrule
  \end{tabular}
\end{table}

We also test the performance in the four datasets described in Sec.~\ref{sec: datasets} following the traditional protocol.
As shown in Tables~\ref{tbl: trad small}~and~\ref{tbl: trad large},
our mean field losses outperform \mbox{ProxyAnchor} loss in MAP@R except for the Cars dataset,
which is consistent with the first experiment.
The improvement in accuracy is evident in the larger datasets.
Besides, \mbox{MFCont}. and \mbox{MFCWMS} losses converge faster than \mbox{ProxyAnchor} loss in all the datasets.
See also the supplement for plots of learning curves.

\begin{table}[t]
  \caption{
    MAP@R values and epochs with the best accuracies obtained using the traditional protocol~\cite{movshovitz2017no,kim2020proxy} on the CUB and Cars datasets.
    The best result in each block is highlighted in bold.
  }
  \label{tbl: trad small}
  \centering
  \begin{tabular}{c cccc}
    \toprule
               & \multicolumn{2}{c}{CUB} & \multicolumn{2}{c}{Cars}                                                    \\
    \cmidrule(r){2-3}\cmidrule(r){4-5}
    Loss       & MAP@R ($\uparrow$)      & Epoch ($\downarrow$)     & MAP@R ($\uparrow$)      & Epoch ($\downarrow$)   \\
    \midrule
    ProxyAnch. & ${25.1 \pm 0.2}$        & ${11.5 \pm 1.4}$         & $\better{26.3 \pm 0.2}$ & ${23.6 \pm 1.8}$       \\
    \MFCont.   & $\better{25.3 \pm 0.3}$ & $\better{4.4 \pm 0.4}$   & ${24.7 \pm 0.2}$        & ${10.1 \pm 0.5}$       \\
    \MFCWMS    & $\better{25.3 \pm 0.3}$ & ${4.7 \pm 0.6}$          & ${24.0 \pm 0.2}$        & $\better{8.5 \pm 0.8}$ \\
    \bottomrule
    \\
  \end{tabular}
  \caption{
    MAP@R values and epochs with the best accuracies obtained using the traditional protocol~\cite{movshovitz2017no,kim2020proxy} on the SOP and InShop datasets.
    The best result in each block is highlighted in bold.
  }
  \label{tbl: trad large}
  \centering
  \begin{tabular}{c cccc}
    \toprule
               & \multicolumn{2}{c}{SOP} & \multicolumn{2}{c}{InShop}                                                     \\
    \cmidrule(r){2-3}\cmidrule(r){4-5}
    Loss       & MAP@R ($\uparrow$)      & Epoch ($\downarrow$)       & MAP@R ($\uparrow$)      & Epoch ($\downarrow$)    \\
    \midrule
    ProxyAnch. & ${51.5 \pm 0.3}$        & ${40.5 \pm 6.5}$           & ${65.5 \pm 0.1}$        & ${31.3 \pm 7.2}$        \\
    \MFCont.   & $\better{52.9 \pm 0.1}$ & ${25.1 \pm 1.4}$           & $\better{67.7 \pm 0.2}$ & ${24.1 \pm 2.7}$        \\
    \MFCWMS    & ${52.7 \pm 0.0}$        & $\better{23.0 \pm 1.3}$    & ${67.5 \pm 0.4}$        & $\better{20.6 \pm 2.1}$ \\
    \bottomrule
  \end{tabular}
\end{table}

\subsection{Impact of hyperparameters}\label{sec: hyp}

\begin{figure}[t]
  \begin{tabular}{cc}
    \begin{minipage}[t]{0.45\hsize}
      \centering
      \includegraphics[scale=0.24]{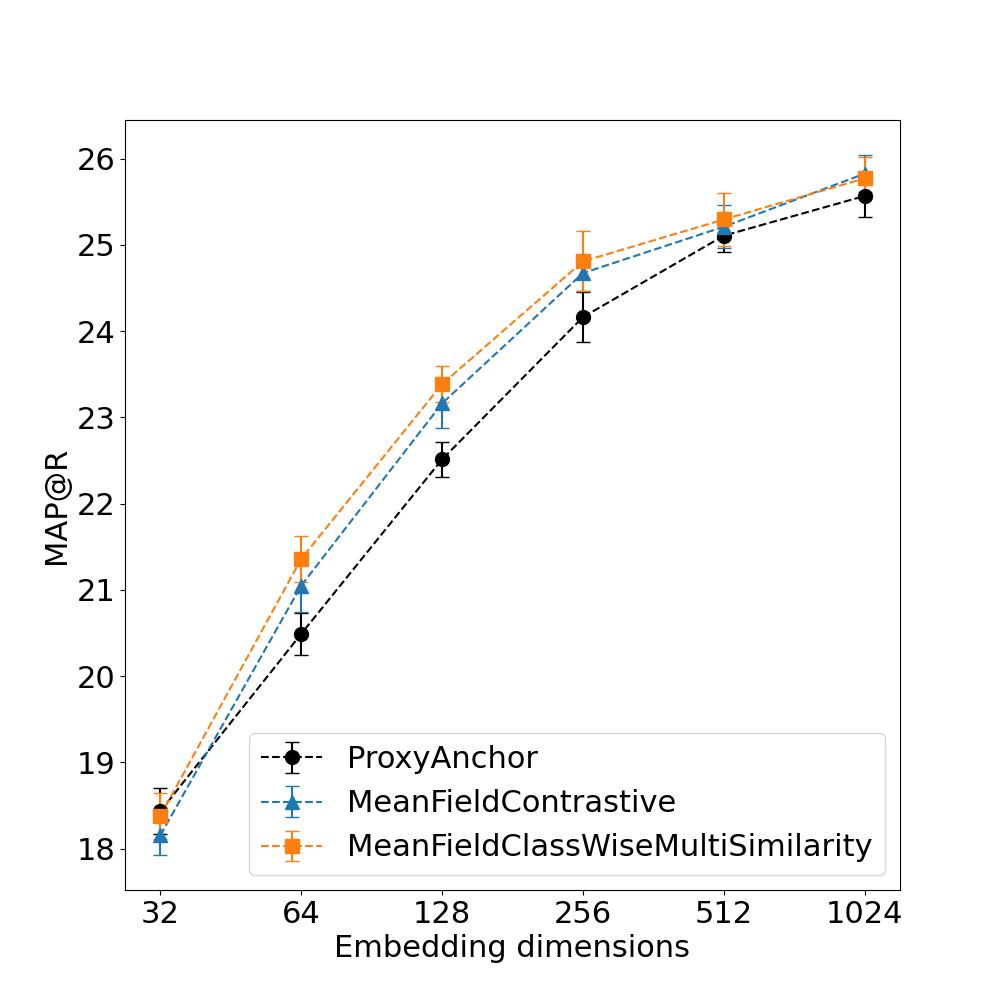}
      \caption{
        MAP@R versus embedding dimensions in CUB,
        comparing \mbox{ProxyAnchor}, \mbox{MFCont}., and \mbox{MFCWMS} losses.
      }\label{fig: emb dim}
    \end{minipage} &
    \begin{minipage}[t]{0.45\hsize}
      \centering
      \includegraphics[scale=0.24]{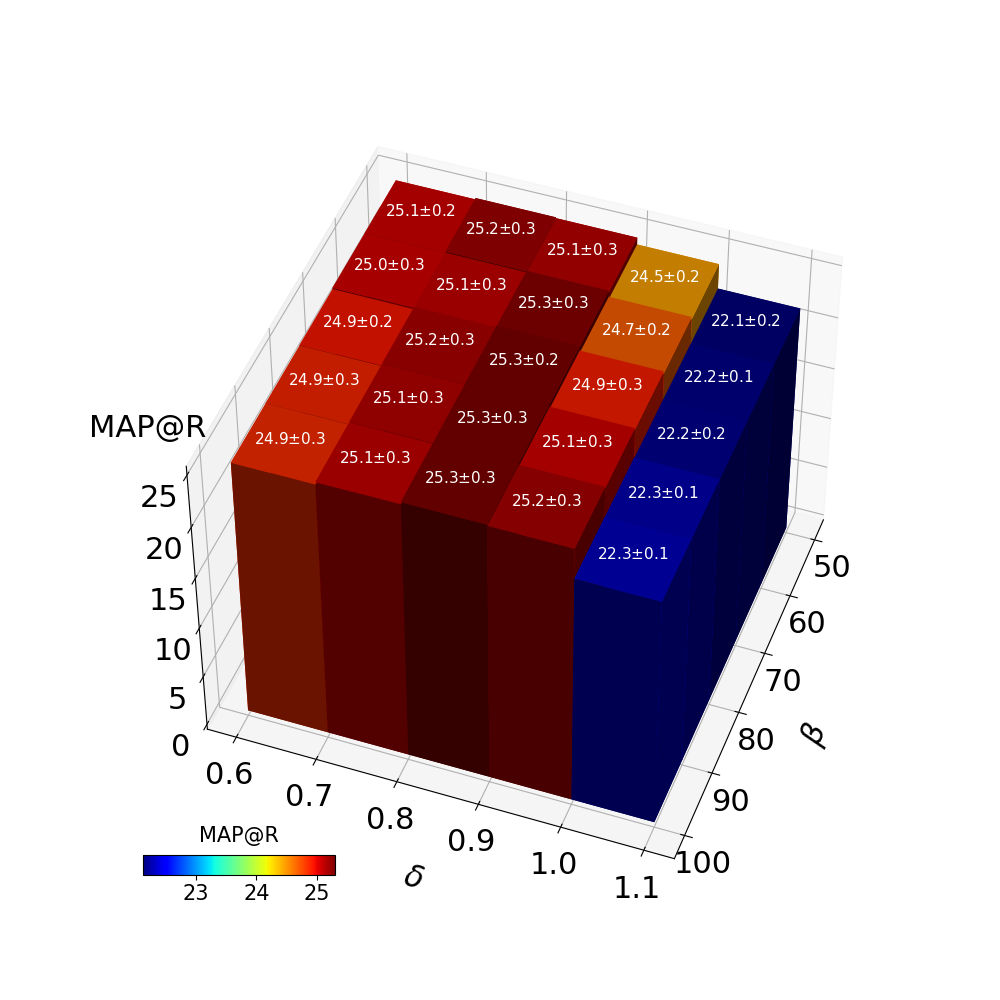}
      \caption{
        MAP@R against $\beta$ and $\delta$ of \mbox{MFCWMS} loss in the CUB dataset.
      }\label{fig: param}
    \end{minipage}
  \end{tabular}
\end{figure}

\paragraph{Embedding dimensions.}
Since embedding dimensions are crucial hyperparameters controlling the performance of image retrieval,
we investigate their effect on the accuracy (MAP@R).
On the CUB dataset, we run the traditional experiments
for \mbox{ProxyAnchor}, \mbox{MFCont}., and \mbox{MFCWMS} losses
by varying the embedding dimensions from $32$ to $1024$.
The results are shown in Fig.~\ref{fig: emb dim}.
We find that our mean field losses monotonically increase their performance
and consistently show better performance compared to \mbox{ProxyAnchor} loss.
Notably, the performance improvement is significantly larger in the cases of $64$, $128$, and $256$ dimensions,
suggesting that our proposed methods potentially offer efficient embeddings in relatively small dimensions.

\paragraph{Batch size.}
We also investigate how batch size affects the accuracy of our models.
We compare the test performance of \mbox{MFCWMS} loss by changing the batch size
between $30$ and $180$ for CUB and Cars (Table~\ref{tbl: batch small}) and $30$ and $300$ for SOP and InShop (Table~\ref{tbl: batch large}).
The accuracy metric basically improves as the batch size increases.
The improvement mostly saturates around $150$ and $180$ for the smaller datasets
and around $150$ and $300$ for SOP.
In the InShop dataset, we observe that the accuracy drops around batch size $150$ with a relatively large variance,
but this can be consistent with \mbox{ProxyAnchor} loss since it also decreases its accuracy around this batch size
(see the supplement).

\begin{table}[t]
  \begin{tabular}{cc}
    \begin{minipage}[t]{0.45\hsize}
      \centering
      \caption{Impact of batch size to MAP@R for \mbox{MFCWMS} loss in CUB and Cars.}\label{tbl: batch small}
      \begin{tabular}{ccc}
        \toprule
        Batch size & CUB                     & Cars                    \\
        \midrule
        ${30}$     & ${23.3 \pm 0.3}$        & ${22.0 \pm 0.3}$        \\
        ${60}$     & ${24.4 \pm 0.1}$        & ${23.2 \pm 0.3}$        \\
        ${90}$     & ${25.1 \pm 0.1}$        & ${23.6 \pm 0.4}$        \\
        ${120}$    & ${25.3 \pm 0.2}$        & ${23.9 \pm 0.2}$        \\
        ${150}$    & $\better{25.4 \pm 0.3}$ & $\better{24.2 \pm 0.3}$ \\
        ${180}$    & $\better{25.4 \pm 0.2}$ & $\better{24.2 \pm 0.2}$ \\
        \bottomrule
      \end{tabular}
    \end{minipage} &
    \begin{minipage}[t]{0.45\hsize}
      \centering
      \caption{Impact of batch size to MAP@R for \mbox{MFCWMS} loss in SOP and InShop.}\label{tbl: batch large}
      \begin{tabular}{ccc}
        \toprule
        Batch size & SOP                     & InShop                  \\
        \midrule
        ${30}$     & ${51.1 \pm 0.1}$        & ${67.4 \pm 0.2}$        \\
        ${60}$     & ${52.0 \pm 0.1}$        & ${67.6 \pm 0.2}$        \\
        ${90}$     & ${52.4 \pm 0.1}$        & ${67.6 \pm 0.2}$        \\
        ${120}$    & ${52.7 \pm 0.1}$        & $\better{67.8 \pm 0.2}$ \\
        ${150}$    & ${52.8 \pm 0.1}$        & ${67.0 \pm 0.6}$        \\
        ${300}$    & $\better{53.0 \pm 0.3}$ & ${67.0 \pm 0.4}$        \\
        \bottomrule
      \end{tabular}
    \end{minipage}
  \end{tabular}
\end{table}

\paragraph{$\beta$ and $\delta$ of \mbox{MFCWMS}.}

We also explore the effect of $\beta$ and $\delta$ of \mbox{MFCWMS} loss in the CUB dataset.
We varied $\beta$ from $50$ to $90$ and $\delta$ from $0.6$ to $1$,
fixing $(\alpha,\lambda_\mrm{MF})$ to $(0.01,0)$ and computed the MAP@R for the test data.
The results are summarized in Fig.~\ref{fig: param}.
Fig.~\ref{fig: param} ensures the competitive performance of the \mbox{MFCWMS} loss is stable against a choice of the hyperparameters.
We observe that the preferred $\beta$ gradually decreases as the margin parameter $\delta$ decreases.

\paragraph{Mean field regularization.}

Finally, we study the impact of the regularization term in Eq.~\eqref{eq: mfcwms}.
We investigate the test performance of \mbox{MFCont}. and \mbox{MFCWMS} losses in the CUB dataset
varying the scale of the regularization coefficient, which is summarized in Table~\ref{tbl: reg}.
The results indicate that the regularization does not offer a statistically significant difference in performance.
This is because the constraint~\eqref{eq: cont constraint} should be satisfied naturally
when the main parts of the mean field losses are minimized.
Thus, we conclude that the constraints in \mbox{MFCont}. and \mbox{MFCWMS} losses are significant theoretically,
but less relevant in practice.

\begin{table}[t]
  \caption{MAP@R against the regularization coefficient $\lambda_\mrm{MF}$ in Eq.~\eqref{eq: mfcwms} in the CUB dataset.}
  \label{tbl: reg}
  \begin{tabular}{ccccccc}
    \toprule
    $\lambda_\mrm{MF}$ & 0                       & 0.01                    & 0.1                     & 1                       & 10               & 100              \\
    \midrule
    \MFCont.           & ${25.2 \pm 0.3}$        & ${25.2 \pm 0.3}$        & $\better{25.3 \pm 0.3}$ & ${25.2 \pm 0.2}$        & ${25.2 \pm 0.3}$ & ${25.0 \pm 0.3}$ \\
    \MFCWMS            & $\better{25.3 \pm 0.2}$ & $\better{25.3 \pm 0.3}$ & $\better{25.3 \pm 0.3}$ & $\better{25.3 \pm 0.3}$ & ${25.2 \pm 0.3}$ & ${25.1 \pm 0.3}$ \\
    \bottomrule
  \end{tabular}
\end{table}

\section{Conclusion}

In this paper, we explored applications of the mean field theory in statistical physics
to address the challenges in training complexity associated with conventional metric learning loss functions.
The mean field theory replaces interactions among data points with interactions with mean fields,
approximating pair-based losses by classification-based ones.
While the mean field theory shares a similar philosophy with the proxy-based method~\cite{movshovitz2017no},
it relies on expansions with respect to fluctuations around the mean fields
and is suitable for applications to pair-based losses without anchors.

Based on the analogy between deep metric learning and magnets,
we applied the mean field theory to Contrastive loss and \mbox{ClassWiseMultiSimilarity} loss,
a variant of MultiSimilarity loss~\cite{wang2019multi} without anchors,
and derived \mbox{MeanFieldContrastive} and \mbox{MeanFieldClassWiseMultiSimilarity} losses.
We extensively evaluated the proposed loss functions
and compared them with the existing baseline methods using both modern and traditional benchmark protocols.
The evaluation results demonstrate that the proposed loss functions outperform the baselines
in the CUB-200-2011 (CUB) and Stanford Online Products (SOP) datasets in the former protocol,
and in the CUB, SOP, and InShop datasets in the latter.
These findings highlight the potential of mean field theory as a powerful tool for simplifying
and improving deep metric learning performance in various machine learning applications.

\section*{Limitations}
Despite the promising results demonstrated by the proposed loss functions,
this work has the following limitations that should be acknowledged:
\Item{
  \item Theoretical guarantees:
  Since our approach is based on the idea of physics,
  our discussion primarily focuses on its conceptual description and practical utility.
  Its mathematical aspects, including rigorous proof and theoretical analysis of convergence properties, have not been explored.
  \item Multi-label settings:
  The current study focuses on single-label classification tasks, where each image belongs to only one class.
  The applicability and performance of the proposed loss functions in multi-label settings, where an image can belong to multiple classes, have not been investigated.
}
These limitations should be addressed in future work.

\begin{ack}
  We thank Yuki Saito, Ryosuke Goto, and Masanari Kimura for their useful comments on our manuscript.
\end{ack}

\bibliographystyle{unsrtnat}
\bibliography{mft-in-dml.bib}

\newpage

\newgeometry{left=3cm,top=2cm,bottom=2cm}

\vspace*{-1cm}
\noindent\rule{\textwidth}{1pt}
\begin{center}
  \raisebox{-0.5\height}{\fontsize{20}{24}\selectfont{Supplementary Material}}
\end{center}
\noindent\rule{\textwidth}{1pt}
\vspace{1cm}

\appendix

\section{Additional experimental results}

In this section, we present the experimental results that cannot be shown in the main paper due to the page limit.

\paragraph{MLRC results.}
Tables~\ref{tbl: cub full}~--~\ref{tbl: sop full} show the complete results of Table~1 in the main paper,
which are obtained the modern benchmark protocol proposed in the ``Metric Learning Reality Check'' (MLRC) paper~\cite{musgrave2020metric}.
In the CUB-200-2011 (CUB) dataset~\cite{wah2011caltech},
MeanFieldContrastive (MFCont.) and MeanFieldClassWiseMultiSimilarity (MFCWMS) losses
outperform the others in Mean Average Precision at R (MAP@R) and R-Precision (RP),
while ProxyAnchor loss~\cite{kim2020proxy} is better in Precision at $1$ (P@1) in the separated case.
In contrast, in the Stanford Online Products (SOP) dataset~\cite{oh2016deep},
the MFCWMS loss shows the best performance in all the metrics.

\begin{figure}[t]
  \begin{tabular}{cc}
    \begin{minipage}[t]{0.45\hsize}
      \centering
      \includegraphics[scale=0.24]{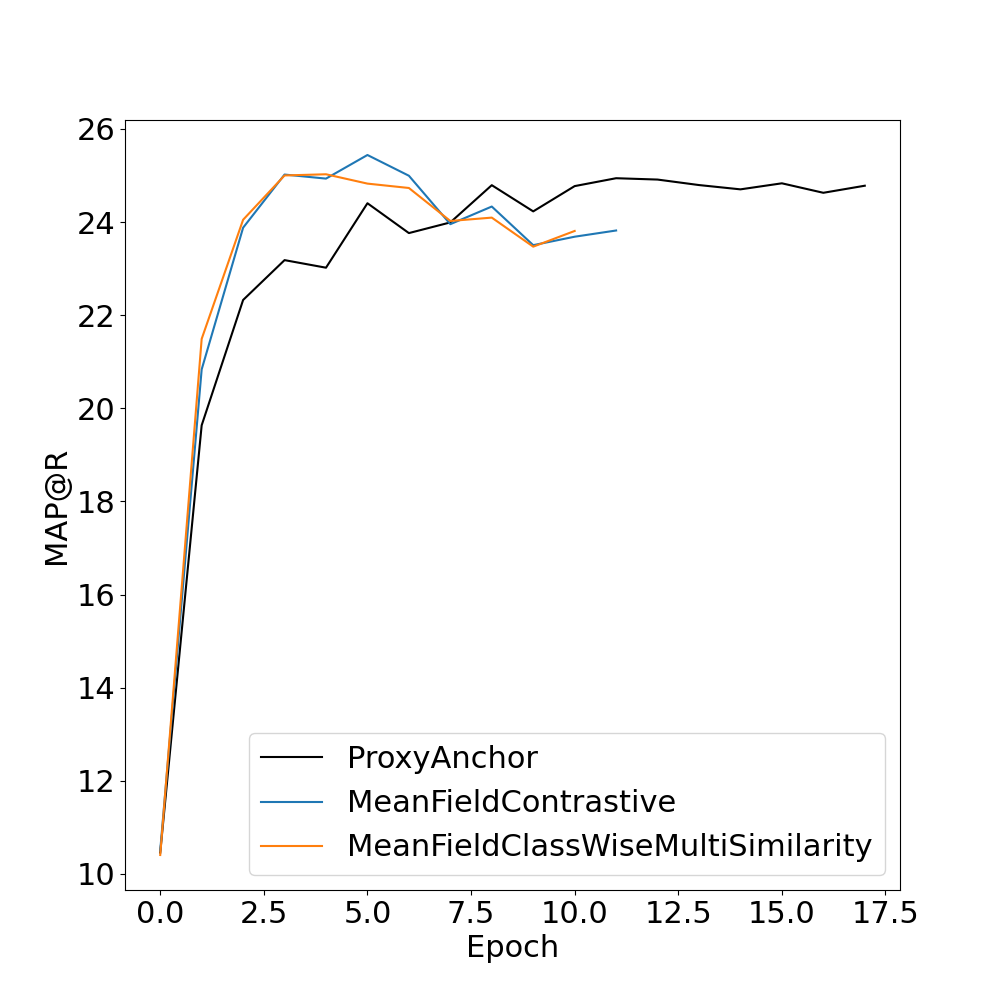}
      \subcaption{CUB}\label{fig: learning curve cub}
    \end{minipage} &
    \begin{minipage}[t]{0.45\hsize}
      \centering
      \includegraphics[scale=0.24]{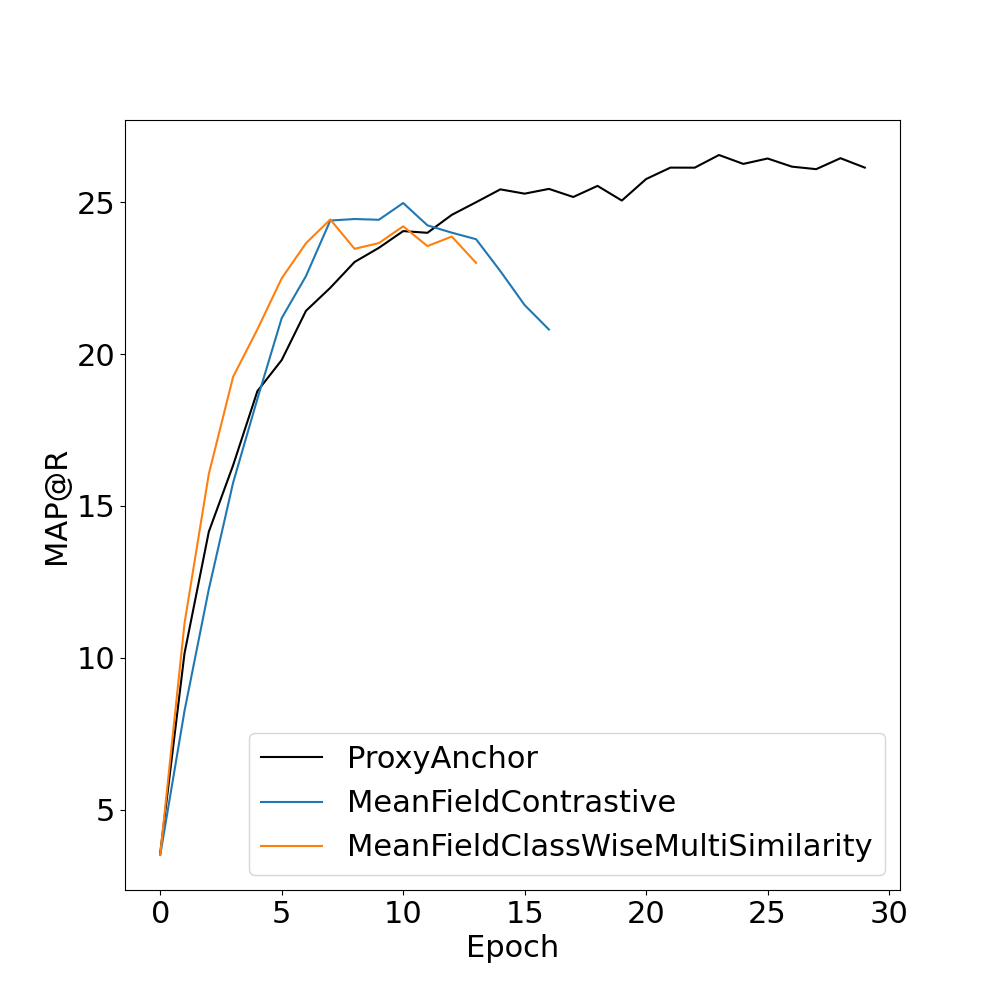}
      \subcaption{Cars}\label{fig: learning curve cars}
    \end{minipage}
  \end{tabular}
  \begin{tabular}{cc}
    \begin{minipage}[t]{0.45\hsize}
      \centering
      \includegraphics[scale=0.24]{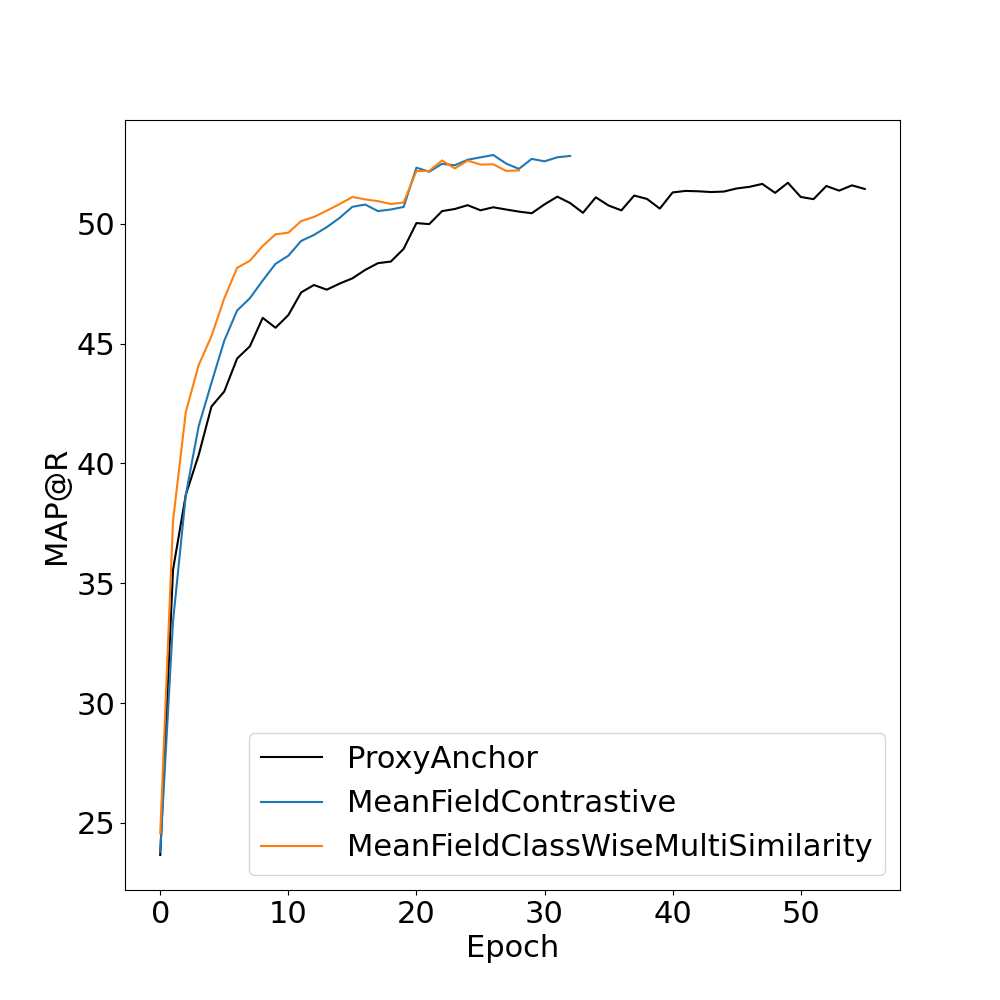}
      \subcaption{SOP}\label{fig: learning curve sop}
    \end{minipage} &
    \begin{minipage}[t]{0.45\hsize}
      \centering
      \includegraphics[scale=0.24]{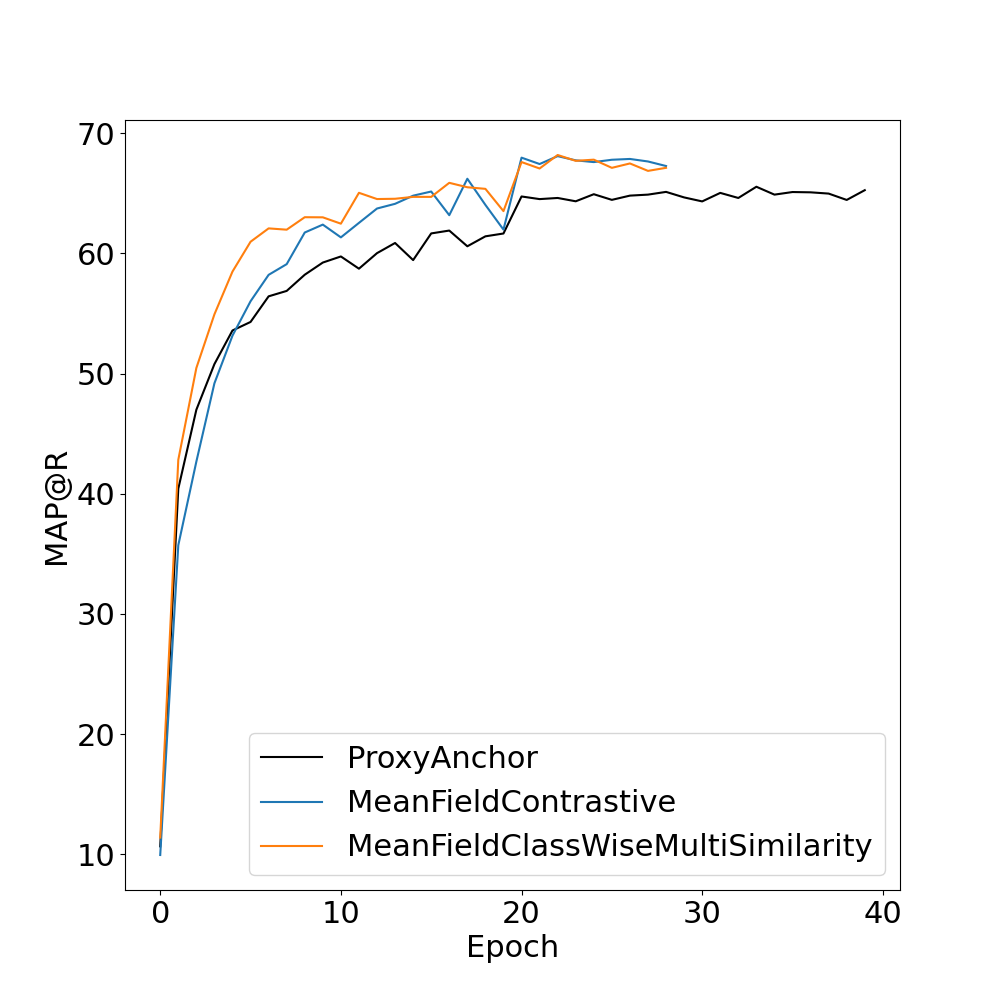}
      \subcaption{InShop}\label{fig: learning curve inshop}
    \end{minipage}
  \end{tabular}
  \caption{
    The test accuracy (MAP@R) plotted against the number of epochs for the (a) CUB, (b) Cars, (c) SOP, and (d) InShop datasets,
    comparing ProxyAnchor, MFCont., and MFCWMS.
  }\label{fig: learning curves}
\end{figure}

\paragraph{Learning curves.}
Figure~\ref{fig: learning curves} shows learning curves obtained
in the traditional evaluation protocol~\cite{movshovitz2017no,kim2020proxy} in fixed seeds,
associated with Tables~2~and~3 in the main paper.
Both MFCont. and MFCWMS losses show faster convergence than the ProxyAnchor loss.
In the smaller datasets (CUB and Cars), accuracies of our mean field losses seem to decrease faster
while we don't see such behaviors in the larger datasets (SOP and InShop~\cite{liu2016deepfashion}).
This phenomenon might be caused by strong repulsive interactions with negative mean fields.
For larger datasets, the embedding spaces may be sufficiently populated to balance the repulsive force,
while this may not be the case for smaller datasets.
It might not occur for ProxyAnchor loss
since repulsive forces for ProxyAnchor loss are weighted
depending on distances between proxy and negative samples.

\paragraph{Impact of batch size in InShop.}
Table~\ref{tbl: batch inshop full} compares the MAP@R in ProxyAnchor and MFCWMS losses
in the InShop dataset varying the batch size.
As mentioned in the main paper, the accuracy of ProxyAnchor starts to decrease gradually for large batch sizes,
while that of MFCWMS loss drops at batch size $150$.
Moreover, Table~\ref{tbl: batch inshop simple split full} shows
the MAP@R in ProxyAnchor and MFCWMS for the InShop dataset without the query--gallery split of test data.
In this case, accuracies of both losses start to decrease gradually around batch size $150$.
Thus, the accuracy drop for the MFCWMS loss in Table~\ref{tbl: batch inshop full} probably comes from the specific query--gallery split.

\tabcolsep=0.15cm
\begin{table}
  \caption{
    MLRC evaluation results in CUB-200-2011~\cite{wah2011caltech}.
    We carry out $10$ test runs and show averaged metrics with their confidence intervals.
  }
  \label{tbl: cub full}
  \centering
  \begin{tabular}{ccccccc}
    \toprule
               & \multicolumn{3}{c}{Separated ($128$D)} & \multicolumn{3}{c}{Concatenated ($512$D)}                                                                                                         \\
    \cmidrule(r){2-4}\cmidrule(r){5-7}
    Loss       & MAP@R                                  & P@1                                       & RP                      & MAP@R                   & P@1                     & RP                      \\
    \midrule
    ArcFace    & ${21.46\pm0.13}$                       & ${59.98\pm0.22}$                          & ${32.31\pm0.14}$        & ${26.39\pm0.16}$        & ${67.11\pm0.23}$        & ${37.23\pm0.17}$        \\
    CosFace    & ${21.19\pm0.22}$                       & ${59.74\pm0.28}$                          & ${32.00\pm0.23}$        & $\better{26.54\pm0.29}$ & ${67.14\pm0.29}$        & $\better{37.38\pm0.28}$ \\
    MS         & ${20.98\pm0.16}$                       & ${59.38\pm0.27}$                          & ${31.84\pm0.15}$        & ${26.20\pm0.16}$        & ${67.34\pm0.35}$        & ${36.99\pm0.16}$        \\
    MS+Miner   & ${20.78\pm0.17}$                       & ${59.02\pm0.25}$                          & ${31.67\pm0.16}$        & ${25.94\pm0.18}$        & ${67.08\pm0.32}$        & ${36.77\pm0.16}$        \\
    ProxyNCA   & ${18.75\pm0.18}$                       & ${57.06\pm0.27}$                          & ${29.64\pm0.21}$        & ${23.84\pm0.22}$        & ${65.60\pm0.28}$        & ${34.82\pm0.25}$        \\
    ProxyAnch. & $\better{21.67\pm0.22}$                & $\best{60.80\pm0.33}$                     & $\better{32.53\pm0.23}$ & ${26.48\pm0.23}$        & $\better{67.72\pm0.30}$ & ${37.30\pm0.23}$        \\
    \midrule
    Cont.      & ${21.02\pm0.14}$                       & ${59.35\pm0.33}$                          & ${31.80\pm0.15}$        & ${26.37\pm0.18}$        & $\better{67.67\pm0.25}$ & ${37.10\pm0.19}$        \\
    MFCont.    & $\better{22.01\pm0.10}$                & $\better{60.29\pm0.23}$                   & $\better{32.85\pm0.10}$ & $\best{27.16\pm0.07}$   & ${67.64\pm0.27}$        & $\best{37.95\pm0.07}$   \\
    \midrule
    CWMS       & ${21.48\pm0.27}$                       & ${60.09\pm0.27}$                          & ${32.32\pm0.26}$        & ${26.94\pm0.29}$        & $\best{68.24\pm0.42}$   & ${37.69\pm0.27}$        \\
    MFCWMS     & $\best{22.11\pm0.08}$                  & $\better{60.28\pm0.10}$                   & $\best{32.96\pm0.08}$   & $\better{27.03\pm0.12}$ & ${67.63\pm0.21}$        & $\better{37.83\pm0.12}$ \\
    \bottomrule
    \\
    \\
  \end{tabular}
  \caption{MLRC evaluation results in Cars-196~\cite{krause20133d}.
  We carry out $10$ test runs and show averaged metrics with their confidence intervals.
  }
  \label{tbl: cars full}
  \centering
  \begin{tabular}{ccccccc}
    \toprule
               & \multicolumn{3}{c}{Separated ($128$D)} & \multicolumn{3}{c}{Concatenated ($512$D)}                                                                                                         \\
    \cmidrule(r){2-4}\cmidrule(r){5-7}
    Loss       & MAP@R                                  & P@1                                       & RP                      & MAP@R                   & P@1                     & RP                      \\
    \midrule
    ArcFace    & ${18.25\pm0.12}$                       & ${71.12\pm0.36}$                          & ${28.63\pm0.13}$        & $\better{27.63\pm0.15}$ & ${84.39\pm0.15}$        & $\better{37.45\pm0.15}$ \\
    CosFace    & ${18.49\pm0.13}$                       & ${74.66\pm0.21}$                          & ${28.75\pm0.12}$        & ${26.96\pm0.25}$        & ${85.29\pm0.26}$        & ${36.80\pm0.24}$        \\
    MS         & ${18.66\pm0.30}$                       & ${71.89\pm0.33}$                          & ${29.42\pm0.29}$        & ${27.19\pm0.41}$        & ${84.03\pm0.30}$        & ${37.39\pm0.36}$        \\
    MS+Miner   & ${18.49\pm0.23}$                       & ${71.99\pm0.28}$                          & ${29.20\pm0.23}$        & ${26.89\pm0.38}$        & ${83.89\pm0.36}$        & ${37.09\pm0.33}$        \\
    ProxyNCA   & ${17.43\pm0.11}$                       & ${70.96\pm0.26}$                          & ${27.85\pm0.10}$        & ${26.78\pm0.18}$        & ${84.31\pm0.24}$        & ${36.83\pm0.17}$        \\
    ProxyAnch. & $\best{19.44\pm0.17}$                  & $\best{76.15\pm0.25}$                     & $\better{29.89\pm0.18}$ & ${26.81\pm0.27}$        & $\best{85.53\pm0.30}$   & ${36.76\pm0.26}$        \\
    \midrule
    Cont.      & ${17.04\pm0.26}$                       & ${69.77\pm0.40}$                          & ${27.48\pm0.26}$        & ${24.93\pm0.46}$        & ${81.87\pm0.35}$        & ${35.12\pm0.42}$        \\
    MFCont.    & $\better{18.12\pm0.13}$                & $\better{71.77\pm0.28}$                   & $\better{28.54\pm0.14}$ & $\better{27.37\pm0.18}$ & $\better{84.56\pm0.21}$ & $\better{37.19\pm0.18}$ \\
    \midrule
    CWMS       & $\better{19.27\pm0.26}$                & $\better{74.19\pm0.30}$                   & $\best{29.95\pm0.25}$   & $\best{27.80\pm0.33}$   & $\better{85.18\pm0.28}$ & $\best{37.89\pm0.29}$   \\
    MFCWMS     & ${18.85\pm0.16}$                       & ${73.02\pm0.20}$                          & ${29.55\pm0.15}$        & ${26.98\pm0.31}$        & ${84.00\pm0.22}$        & ${37.11\pm0.27}$        \\
    \bottomrule
    \\
    \\
  \end{tabular}
  \caption{MLRC evaluation results in Stanford Online Products~\cite{oh2016deep}.
  We carry out $10$ test runs and show averaged metrics with their confidence intervals.
  We remove ProxyAnchor because it failed to converge in our settings.
  }
  \label{tbl: sop full}
  \centering
  \begin{tabular}{ccccccc}
    \toprule
             & \multicolumn{3}{c}{Separated ($128$D)} & \multicolumn{3}{c}{Concatenated ($512$D)}                                                                                                         \\
    \cmidrule(r){2-4}\cmidrule(r){5-7}
    Loss     & MAP@R                                  & P@1                                       & RP                      & MAP@R                   & P@1                     & RP                      \\
    \midrule
    ArcFace  & ${41.47\pm0.24}$                       & ${71.39\pm0.20}$                          & ${44.35\pm0.23}$        & $\better{47.37\pm0.23}$ & $\better{76.13\pm0.16}$ & $\better{50.22\pm0.22}$ \\
    CosFace  & ${41.01\pm0.24}$                       & ${71.03\pm0.22}$                          & ${43.89\pm0.24}$        & ${46.77\pm0.20}$        & ${75.69\pm0.13}$        & ${49.63\pm0.20}$        \\
    MS       & ${41.87\pm0.21}$                       & ${71.10\pm0.18}$                          & ${45.00\pm0.20}$        & ${46.70\pm0.18}$        & ${75.21\pm0.15}$        & ${49.70\pm0.17}$        \\
    MS+Miner & ${41.90\pm0.30}$                       & ${71.08\pm0.25}$                          & ${45.05\pm0.30}$        & ${46.57\pm0.28}$        & ${75.09\pm0.19}$        & ${49.57\pm0.28}$        \\
    ProxyNCA & $\better{42.73\pm0.11}$                & $\better{71.77\pm0.08}$                   & $\better{45.72\pm0.11}$ & ${46.73\pm0.13}$        & ${75.24\pm0.10}$        & ${49.61\pm0.13}$        \\
    \midrule
    Cont.    & ${41.09\pm0.18}$                       & ${70.04\pm0.16}$                          & ${44.18\pm0.19}$        & ${45.35\pm0.19}$        & ${73.88\pm0.15}$        & ${48.28\pm0.19}$        \\
    MFCont.  & $\better{43.62\pm0.36}$                & $\better{72.74\pm0.29}$                   & $\better{46.55\pm0.35}$ & $\better{47.01\pm0.21}$ & $\better{75.57\pm0.16}$ & $\better{49.85\pm0.20}$ \\
    \midrule
    CWMS     & ${41.53\pm0.20}$                       & ${70.76\pm0.16}$                          & ${44.50\pm0.21}$        & ${45.13\pm0.16}$        & ${73.99\pm0.11}$        & ${47.99\pm0.16}$        \\
    MFCWMS   & $\best{44.57\pm0.16}$                  & $\best{73.32\pm0.11}$                     & $\best{47.53\pm0.16}$   & $\best{48.33\pm0.18}$   & $\best{76.38\pm0.14}$   & $\best{51.17\pm0.18}$   \\
    \bottomrule
  \end{tabular}
\end{table}

\begin{table}
  \caption{
    Test accuracies on InShop with the test dataset split into queries and galleries.
  }
  \label{tbl: batch inshop full}
  \centering
  \begin{tabular}{c cc}
    \toprule
    Batch size & ProxyAnchor             & MFCWMS                  \\
    \midrule
    ${30}$     & ${63.6 \pm 1.4}$        & ${67.4 \pm 0.2}$        \\
    ${60}$     & $\better{65.7 \pm 0.2}$ & ${67.6 \pm 0.2}$        \\
    ${90}$     & ${65.5 \pm 0.3}$        & ${67.6 \pm 0.2}$        \\
    ${120}$    & ${65.6 \pm 0.3}$        & $\better{67.8 \pm 0.2}$ \\
    ${150}$    & ${65.5 \pm 0.2}$        & ${67.0 \pm 0.6}$        \\
    ${300}$    & ${64.5 \pm 0.2}$        & ${67.0 \pm 0.4}$        \\
    ${500}$    & ${63.3 \pm 0.2}$        & ${67.1 \pm 0.1}$        \\
    \bottomrule
  \end{tabular}
\end{table}

\begin{table}
  \caption{
    Test accuracies on InShop with the test dataset \textit{not} split into queries and galleries.
  }
  \label{tbl: batch inshop simple split full}
  \centering
  \begin{tabular}{c cc}
    \toprule
    Batch size & ProxyAnchor             & MFCWMS                  \\
    \midrule
    ${30}$     & ${61.7 \pm 0.6}$        & ${64.7 \pm 0.1}$        \\
    ${60}$     & ${62.8 \pm 0.4}$        & $\better{65.1 \pm 0.2}$ \\
    ${90}$     & $\better{62.9 \pm 0.3}$ & ${65.0 \pm 0.5}$        \\
    ${120}$    & $\better{62.9 \pm 0.3}$ & ${64.9 \pm 0.5}$        \\
    ${150}$    & ${62.7 \pm 0.1}$        & ${65.0 \pm 0.4}$        \\
    ${300}$    & ${61.9 \pm 0.2}$        & ${64.7 \pm 0.4}$        \\
    ${500}$    & ${60.6 \pm 0.2}$        & ${64.6 \pm 0.1}$        \\
    \bottomrule
  \end{tabular}
\end{table}

\clearpage

\end{document}